\documentclass[10pt,twocolumn,letterpaper]{article}

\usepackage[pagenumbers]{cvpr} % To force page numbers, e.g. for an arXiv version

\usepackage[dvipsnames]{xcolor}

\definecolor{cvprblue}{rgb}{0.21,0.49,0.74}
\usepackage[breaklinks,colorlinks,citecolor=cvprblue]{hyperref}

\usepackage{multirow}
\usepackage{makecell}

\def\paperID{7028} % *** Enter the Paper ID here
\def\confName{CVPR}
\def\confYear{2024}

\title{A\&B BNN: Add\&Bit-Operation-Only Hardware-Friendly Binary Neural Network}

\author{Ruichen Ma, Guanchao Qiao, Yian Liu, Liwei Meng, Ning Ning, Yang Liu, Shaogang Hu\thanks{Corresponding author}\\
University of Electronic Science and Technology of China\\
{\tt\small ruichen.ma@std.uestc.edu.cn, sghu@uestc.edu.cn}
}

\begin{document}
\maketitle

%% =============================================================== Abstract =============================================================== %% 
%\input{sec/0_abstract}
\begin{abstract}
Binary neural networks utilize 1-bit quantized weights and activations to reduce both the model's storage demands and computational burden.
However, advanced binary architectures still incorporate millions of inefficient and non-hardware-friendly full-precision multiplication operations.
A\&B BNN is proposed to directly remove part of the multiplication operations in a traditional BNN and replace the rest with an equal number of bit operations, introducing the mask layer and the quantized RPReLU structure based on the normalizer-free network architecture.
%A\&B BNN is proposed to replace traditional BNN's multiplication operations with add and bit operations, introducing the mask layer and the quantized RPReLU structure based on the normalizer-free network architecture.
The mask layer can be removed during inference by leveraging the intrinsic characteristics of BNN with straightforward mathematical transformations to avoid the associated multiplication operations.
The quantized RPReLU structure enables more efficient bit operations by constraining its slope to be integer powers of 2.
Experimental results achieved 92.30\%, 69.35\%, and 66.89\% on the CIFAR-10, CIFAR-100, and ImageNet datasets, respectively, which are competitive with the state-of-the-art.
Ablation studies have verified the efficacy of the quantized RPReLU structure, leading to a 1.14\% enhancement on the ImageNet compared to using a fixed slope RLeakyReLU.
The proposed add\&bit-operation-only BNN offers an innovative approach for hardware-friendly network architecture.
\end{abstract}

%% =============================================================== Introduction =============================================================== %% 
%\input{sec/1_intro}
\section{Introduction}
\label{sec:Introduction}

Neural networks have made remarkable strides in tasks including image classification \cite{krizhevsky2012imagenet, simonyan2015very, he2016deep}, object detection \cite{redmon2016you, zhu2021tph}, speech recognition \cite{zhang2020transformer, chan2016listen}, and text generation \cite{radford2018improving, radford2019language, brown2020language}, significantly advancing the development of various fields.
However, as the scale of deep neural networks (DNNs) expands, the substantial computational and storage requirements make them feasible only for running on powerful but expensive GPUs, leaving edge devices unattainable \cite{deng2020model, hung2021four}.
Hardware-efficient network architectures, such as spiking neural networks (SNNs), although they may not outperform traditional networks in numerous domains, offer the advantage of eliminating multiplication operations \cite{tavanaei2019deep, taherkhani2020review}.
This reduction in hardware complexity substantially reduces expenses associated with chip design, making them appealing to chip designers \cite{roy2019towards, zhang2020neuro, mehonic2022brain}.

Numerous approaches have been devised to mitigate hardware overhead in traditional DNNs, albeit often at the expense of modest performance compromises.
Binary neural networks (BNNs) stand out among these approaches, aiming to achieve 1-bit quantization of network parameters and activations to reduce storage and computational requirements.
Pioneering study \cite{courbariaux2016binarized} has enabled BNNs to perform inference exclusively through logical operations, leading to a notable decrease in chip power consumption and design cost.
Recognizing its suboptimal performance on large datasets such as ImageNet, subsequent studies have introduced various techniques to improve overall performance.
These techniques have become nearly indispensable for advanced BNN models, including scaling factors \cite{rastegari2016xnor}, BN \cite{ioffe2015batch} layers, PReLU \cite{he2015delving} layers, and real-value residuals \cite{liu2018bi}.
However, these layers will unavoidably introduce full-precision multiplication operations that are not conducive to hardware efficiency, conflicting with the fundamental goal of BNN.
Although the multiplication operand (MO) introduced is only in the order of millions, it still imposes a significant burden on the hardware, and chip designs that circumvent multipliers are preferable.
\begin{figure*}[!t]
\centering
\includegraphics[width=6.5in]{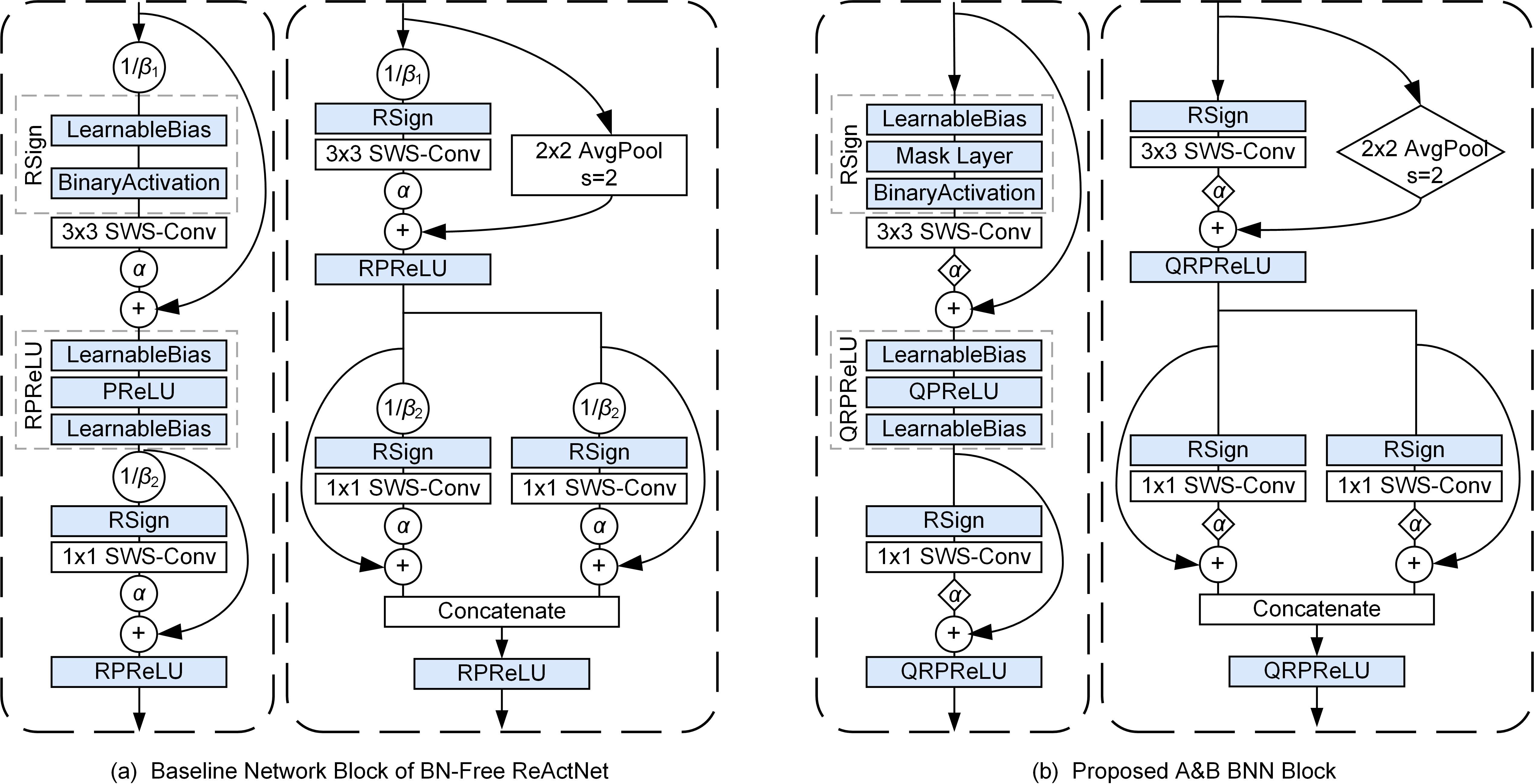}
\caption{The architecture overview of the (a) baseline BN-Free network and (b) proposed A\&B BNN. In contrast to the baseline network, the proposed A\&B BNN eliminates all multiplication operations. The multiplication resulting from $\beta$ is absorbed into the newly introduced mask layer and can be removed directly during inference. Multiplications induced by both average pooling and $\alpha$ are substituted by equal but more efficient bit operations. Additionally, we introduce the quantized RPReLU structure, effectively removing the multiplication associated with PReLU. Circles represent multiplication operations, diamonds represent bit operations, and $\bigoplus$ represents residual addition.}
\label{fig_all}
\end{figure*}
\begin{table}[!t]\small
\centering
\begin{tabular}{p{12.0em} p{4.8em}<{\centering} p{3.5em}<{\centering}}
\toprule[1.5pt]
{Binary Network Architecture} & {MO} & {Acc}\\
\midrule[1pt]
BNN-ResNet-18 \cite{kim2020learning} & 1.51 M & 42.2\% \\
XNOR-ResNet-18 \cite{rastegari2016xnor} & 3.20 M & 51.2\% \\
Bi-ResNet-18 \cite{liu2018bi} & 20.86 M & 56.4\% \\
ReActNet-18 \cite{liu2020reactnet} & 22.37 M & 65.5\% \\
ReActNet-18 (BN-Free) \cite{chen2021bnn} & 4.60 M & 61.1\% \\
Bi-ResNet-34 \cite{liu2018bi} & 22.20 M & 62.2\% \\
ReActNet-A \cite{liu2020reactnet} & 10.79 M & 69.4\% \\
ReActNet-A (BN-Free) \cite{chen2021bnn} & 14.65 M & 68.0\% \\
\midrule[1pt]
\textbf{Bi-ResNet-18 (A\&B BNN)} & \textbf{0} & \textbf{60.38\%} \\
\textbf{ReActNet-18 (A\&B BNN)} & \textbf{0} & \textbf{61.39\%} \\
\textbf{ReActNet-34 (A\&B BNN)} & \textbf{0} & \textbf{65.19\%} \\
\textbf{ReActNet-A (A\&B BNN)} & \textbf{0} & \textbf{66.89\%} \\
\bottomrule[1.5pt]
\end{tabular}
\caption{Top-1 Accuracies of different BNNs evaluated on ImageNet dataset.}
\label{table_compare}
\end{table}

This study presents A\&B BNN, a binary network architecture designed to eliminate all multiplication operations during inference, and was evaluated on three widely used structures ResNet-18/34 and MobileNet.
The accuracies for various mainstream BNNs and A\&B BNN on the ImageNet dataset are presented in \cref{table_compare}, with corresponding results of 61.39\%, 65.19\%, and 66.89\% for the three structures, respectively.
The key to eliminating multiplication lies in removing the BN layer in the network topology while minimizing the loss.
Studies \cite{brock2020characterizing} and \cite{brock2021high} introduced normalizer-free network architecture and study \cite{chen2021bnn} extended this technique to BNN and proposed BN-Free BNN.
While this topology appears to double the multiplication operand, it can be obviated through characteristics of BNN and several mathematical transformations.
The mask layer is proposed to execute these multiplication operations, serving the purpose of gradient scaling during training, and can be removed during inference.
Furthermore, we introduce the quantized RPReLU structure, replacing the multiplication operations introduced by PReLU with an equal number of bit operations.
Experiments show that the proposed approach achieves competitive performance compared to the state-of-the-art on CIFAR-10, CIFAR-100, and ImageNet datasets while eliminating multiplication operations, representing a valuable trade-off.
Our code is available at \url{https://github.com/Ruichen0424/AB-BNN}.

%% =============================================================== Releted Work =============================================================== %% 
\section{Releted Work}
\label{sec:Releted_Work}

{\bf{Binary neural networks.}}
Binary neural networks binarize weights and activations through the sign function, which renders the backpropagation algorithm ineffective due to its non-differentiable characteristic until \cite{courbariaux2016binarized} proposed the straight-through estimator (STE) technique.
STE passes the gradient of binary values to full-precision values directly to disregard the influence of the sign function in the chain rule.
Study \cite{rastegari2016xnor} proposed using scaling factors to compensate the loss incurred during binarization, ${{{\left\| \mathbf{W} \right\|}_{\ell 1}}}/{n}$ for weights and ${{{\left\| \mathbf{a} \right\|}_{\ell 1}}}/{n}$ for activations.
We retain the weight factor as it can be integrated into the weight matrix while excluding the activation factor to avoid introducing any multiplication operations.
\cite{liu2018bi} suggested employing real values in the residual instead of binary to enhance the expressive capacity of BNN.
Since this would introduce multiplication operations, we did not use this technique.
%\cite{liu2018bi} suggested employing real values in the residual instead of binary to enhance the expressive capacity of BNN, which also introduces multiplication operations thus we did not follow.
To enforce BNNs to learn similar distribution as full-precision networks \cite{rastegari2016xnor, xu2019accurate, bulat2019xnor}, \cite{liu2020reactnet} introduced the ReAct Sign (RSign) and ReAct PReLU (RPReLU) structures which added some bias layers based on the original.
\cite{chen2021bnn} applies the BN-Free topology proposed in \cite{brock2020characterizing, brock2021high} to BNN to eliminate the BN layer, which is the foundation of this work and is shown in \cref{fig_all}.
This topology employs the scaled weight standardization technique to regulate the mean and variance of each layer's activations and proposes the residual blocks of the form $x_{\ell+1}=x_{\ell}+\alpha \cdot f_{\ell}(x_{\ell}/\beta_{\ell})$.
$\alpha$ controls the variance growth rate of the residual block and $\beta$ regulates the activation distribution, constituting the primary source of multiplication in BN-Free BNN.

{\bf{Efforts to eliminate multiplication.}}
In BNN, the seminal work \cite{courbariaux2016binarized} introduced XNOR-counts to substitute full-precision multiplication operations, reducing hardware overhead by over 200 times.
Studies \cite{sakr2018true, liu2019circulant, liu2018bi, qin2020forward, wang2020sparsity} proposed diverse STE functions to enhance training effectiveness, while \cite{ding2019regularizing, rozen2022bimodal, darabi2018bnn+, kim2021improving, li2022equal} employed a range of regularization terms to enhance weights and pre-activation distributions, thereby augmenting network expression.
Additionally, there exist studies \cite{liu2021adam, martinez2019training} focused on enhancing network performance through optimizing training strategies and \cite{shang2022lipschitz, gu2019bayesian, zhou2016incremental} dedicated to refining the loss function.
Regrettably, though none of these techniques introduce multiplication operations, almost all of these studies employ methods that do.
In spiking neural networks \cite{tavanaei2019deep, taherkhani2020review}, activations manifest as discrete spikes i.e. \{0, 1\}, enabling computation solely through addition operations.
\cite{chen2020addernet} proposed AdderNet framework wherein $\ell_1$ distance is employed instead of convolution as a metric for assessing the relationship between features and filters within neural networks, with operations limited to addition and subtraction.
Nevertheless, because of the presence of the BN layer, there are still millions of multiplication operations involved, making it not a true multiplication-operation-free network.
\Cref{table_mocounts} shows the multiplication operands introduced by different techniques.
\begin{table}[!t]\small
\centering
\begin{tabular}{p{11.5em} p{11.0em}<{\centering} }
\toprule[1.5pt]
{Techniques} & {MO} \\
\midrule[1pt]
Scaling factor of activations & $(c+1) \cdot h \cdot w$ \\
BN layer & $c \cdot h \cdot w$ \\
PReLU layer & $c \cdot h \cdot w$ \\
Real-value residual & $c_{in} \cdot c_{out} \cdot h_{out} \cdot w_{out} \cdot k_h \cdot k_w$ \\
Average pooling layer & $c_{out} \cdot h_{out} \cdot w_{out}$ \\
$\alpha$ in BN-Free & $c \cdot h \cdot w$ \\
$\beta$ in BN-Free & $c \cdot h \cdot w$ \\
\bottomrule[1.5pt]
\end{tabular}
\caption{Multiplication operands introduced by different techniques.}
\label{table_mocounts}
\end{table}

%% =============================================================== Method =============================================================== %% 
\section{Method}
\label{sec:Method}

This section initially presents the foundational knowledge, followed by an assessment of the multiplication operands and their distribution within the BN-Free structure.
Subsequently, we explore the removable mask layer and then introduce the bit operation and the quantized RPRuLU unit.
Finally, we discuss the practical hardware benefits.

%% --------------------------------------------------------------------------------------------------------------------------
\subsection{Preliminary}

{\bf{Scaled weight standardization.}}
To address the mean shift and variance explode or vanish on activations resulting from BN removal, the scaled weight standardization (SWS) technique from \cite{brock2020characterizing} was introduced.
Specifically, the weights are scaled as follows:
\begin{equation}
\hat{W}_{i,j}=\gamma\cdot\frac{W_{i,j}-\mu_i}{\sqrt{N}\sigma_i}
\end{equation}
where $\mu_i=(1/N) {\textstyle\sum_{j}W_{i,j}}$, $\sigma_i^2=(1/N) {\textstyle\sum_{j}(W_{i,j}-\mu_i)^2}$, $N$ is the fan-in, and $\hat{W}_{i,j}$ is the corresponding standardized weights. $\gamma$ is related to the activation function and is 1 for the sign function.
SWS technique does not introduce any multiplication operation during inference.

{\bf{Adaptive gradient clipping.}}
Gradient clipping is commonly used to restrict the norm of gradients \cite{pascanu2013difficulty} to maintain training stability \cite{merity2018regularizing}.
\cite{brock2021high} proposed the adaptive gradient clipping (AGC) technique to improve the performance of the normalizer-free network, which can be described as:
\begin{equation}
G_{i}^{l} \rightarrow\left\{\begin{array}{ll}
\lambda \cdot \frac{\left\|W_{i}^{l}\right\|_{F}^*}{\left\|G_{i}^{l}\right\|_{F}} G_{i}^{l}, & \text{if } \frac{\left\|G_{i}^{l}\right\|_{F}}{\left\|W_{i}^{l}\right\|_{F}^{*}}>\lambda \\
G_{i}^{l}, & \text{otherwise}
\end{array}\right.
\end{equation}
where $l$ represents the corresponding layer and $i$ represents the $i_{th}$ row of a matrix. ${\left\|W_{i}^{l}\right\|_{F}^{*}}=\max\{{\left\|W_{i}\right\|_{F}}, \epsilon\}$, $\epsilon=10^{-3}$ and ${\left\| \cdot \right\|_{F}}$ is the Frobenius norm.
The clipping threshold $\lambda$ is a crucial hyperparameter that is usually tuned by grid search.

{\bf{Distillation loss functions.}}
Study \cite{liu2020reactnet} introduced Distillation loss to enforce the similarity of distribution between full-precision networks and binary networks to improve performance.
The loss is calculated as follows:
\begin{equation}
\mathcal{L}_{\text {Dis }}=-\frac{1}{n} \sum_{c} \sum_{i=1}^{n} \rho_{c}^{\mathcal{R}}\left(X_{i}\right) \times \log \left(\frac{\rho_{c}^{\mathcal{B}}\left(X_{i}\right)}{\rho_{c}^{\mathcal{R}}\left(X_{i}\right)}\right)
\end{equation}
where $n$ is the batch size, $c$ represents classes and $X_{i}$ is the input image.
$\rho_{c}^{\mathcal{R}}$ is the softmax output of the full-precision network and $\rho_{c}^{\mathcal{B}}$ represents the corresponding softmax output of the binary network.

%% --------------------------------------------------------------------------------------------------------------------------
\subsection{Multiplication Operations in BN-Free BNN}
\begin{figure}[!t]
\centering
\includegraphics[width=3.0in]{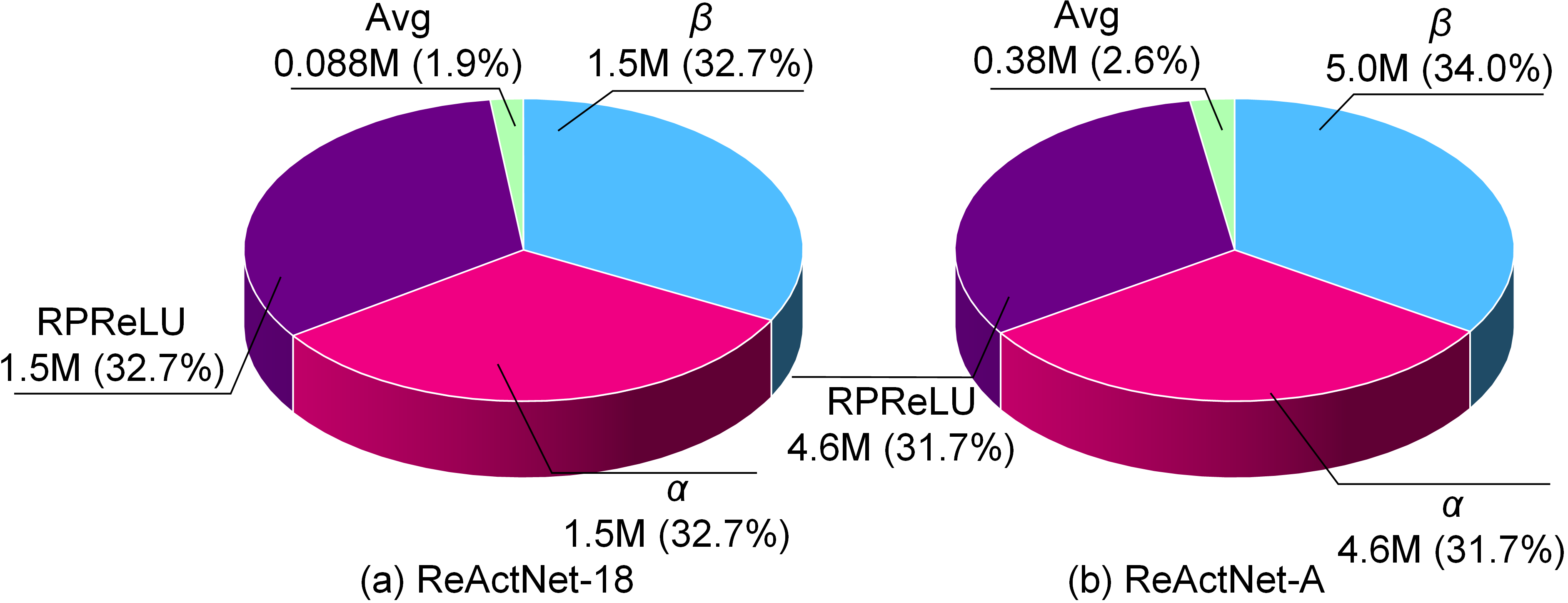}
\caption{The multiplication operand and corresponding ratio within the BN-Free ReActNet-18 and ReActNet-A structures. For input images with a resolution of $224\times224$, the former generates approximately 4.6 million multiplication operations, while the latter yields approximately 14.7 million.}
\label{fig_MO}
\end{figure}

The components that introduce full-precision multiplication operations in the BN-Free BNN architecture \cite{chen2021bnn} include $\alpha$, $\beta$, RPReLU, and the average pooling layer.
$\alpha$ and $\beta$ serve as manually designed scale factors, normalizing the input and output to preserve the advantages of the BN layer.
To normalize the variance, $\beta = \sqrt {{\mathop{\rm Var}\nolimits}({x_{in}})}$ is applied before the convolution operation, and then multiplied by $\alpha$ for further computations.
Typically, $\beta$ corresponds to the expected empirical standard deviation of the activation during initialization, whereas $\alpha$ is commonly assigned a small value, such as 0.2.
\Cref{fig_MO} illustrates the operand for the multiplication operations and the corresponding ratios of BN-Free-based ReActNet-18 and ReActNet-A on an input picture with the resolution of $224\times224$.
The former performs 4.6 million multiplication operations, while the latter performs 14.7 million.

%% --------------------------------------------------------------------------------------------------------------------------
\subsection{Removable Mask Layer}

The mask layer is introduced first to comprehend the gradient transfer through the sign function in BNN. Then we explain the utilization of mathematical transformation to efficiently eliminate the multiplication operation caused by $\beta$ without incurring any additional cost.

\begin{figure}[!t]
\centering
\includegraphics[width=2.6in]{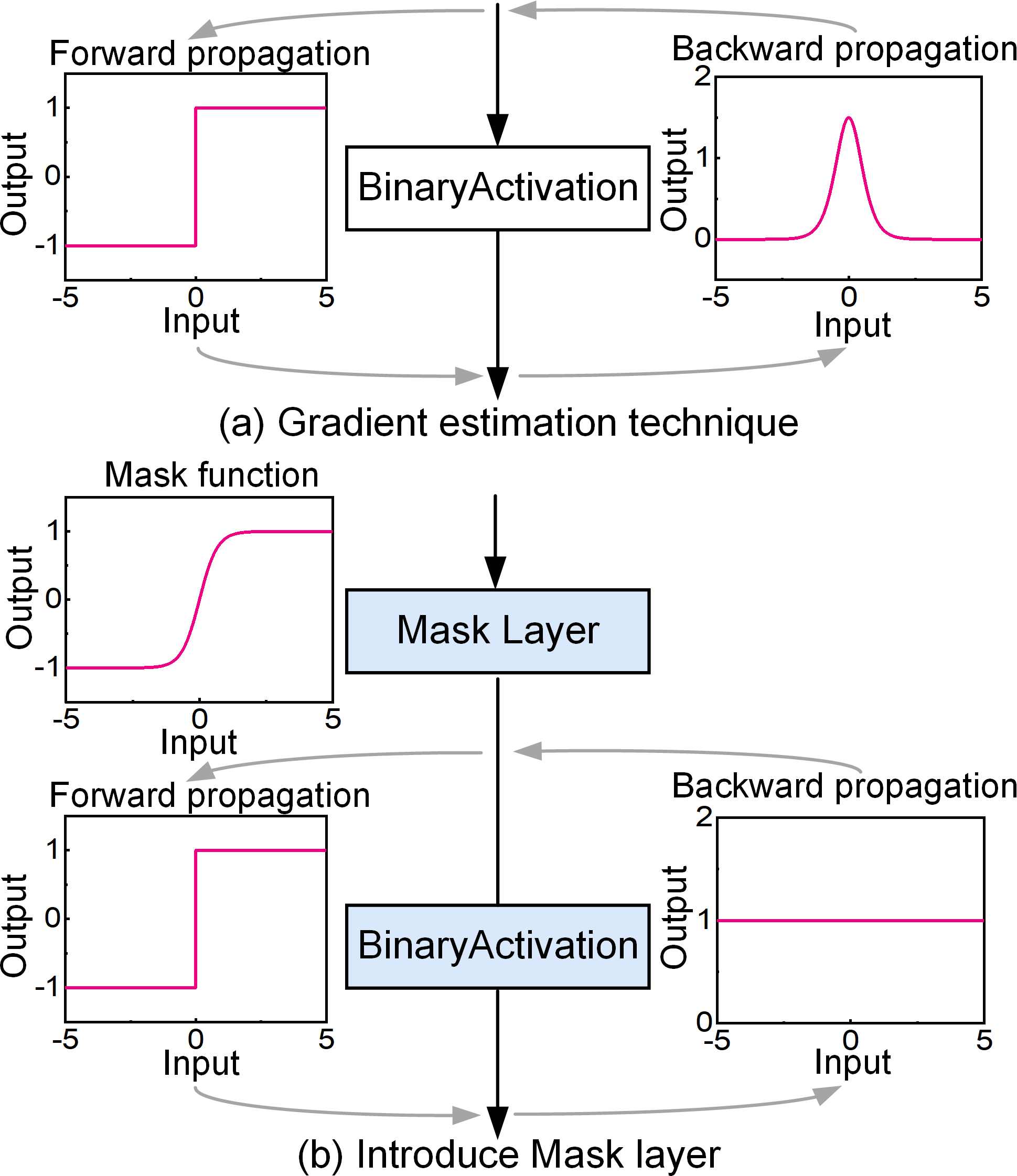}
\caption{(a) A visualization of gradient approximation techniques. The gradient is transferred through an approximate function that resembles the impulse function. (b) Introduce the mask layer to achieve the same effect.}
\label{fig_mask}
\end{figure}

Study \cite{courbariaux2016binarized} proposed the STE technique to realize gradient transfer of the sign function, and subsequent work proposed various approximation functions $f_A(\cdot)$ for optimization. This technique involves using the sign function for binarization during forward propagation and utilizing the derivative of an approximation function during backpropagation to transfer gradients as shown in \cref{fig_mask}a and \cref{equ_BA}.
\begin{equation}
\mathrm{BinaryActivation} \rightarrow\left\{\begin{array}{ll}
\mathrm{Sign}(\cdot), & \text{forward pass} \\
{f'}_A(\cdot), & \text{backward pass}
\end{array}\right.
\label{equ_BA}
\end{equation}
Mathematically, the gradient approximation technique is equivalent to the introduction of a mask layer before the binary activation layer.
This layer serves as an additional activation function, mapping the pre-activations to the binarization layer, and then the binarization layer transfers the gradient during backpropagation directly.
The equivalent process is depicted in \cref{fig_mask}b.

\begin{figure}[!t]
\centering
\includegraphics[width=\linewidth]{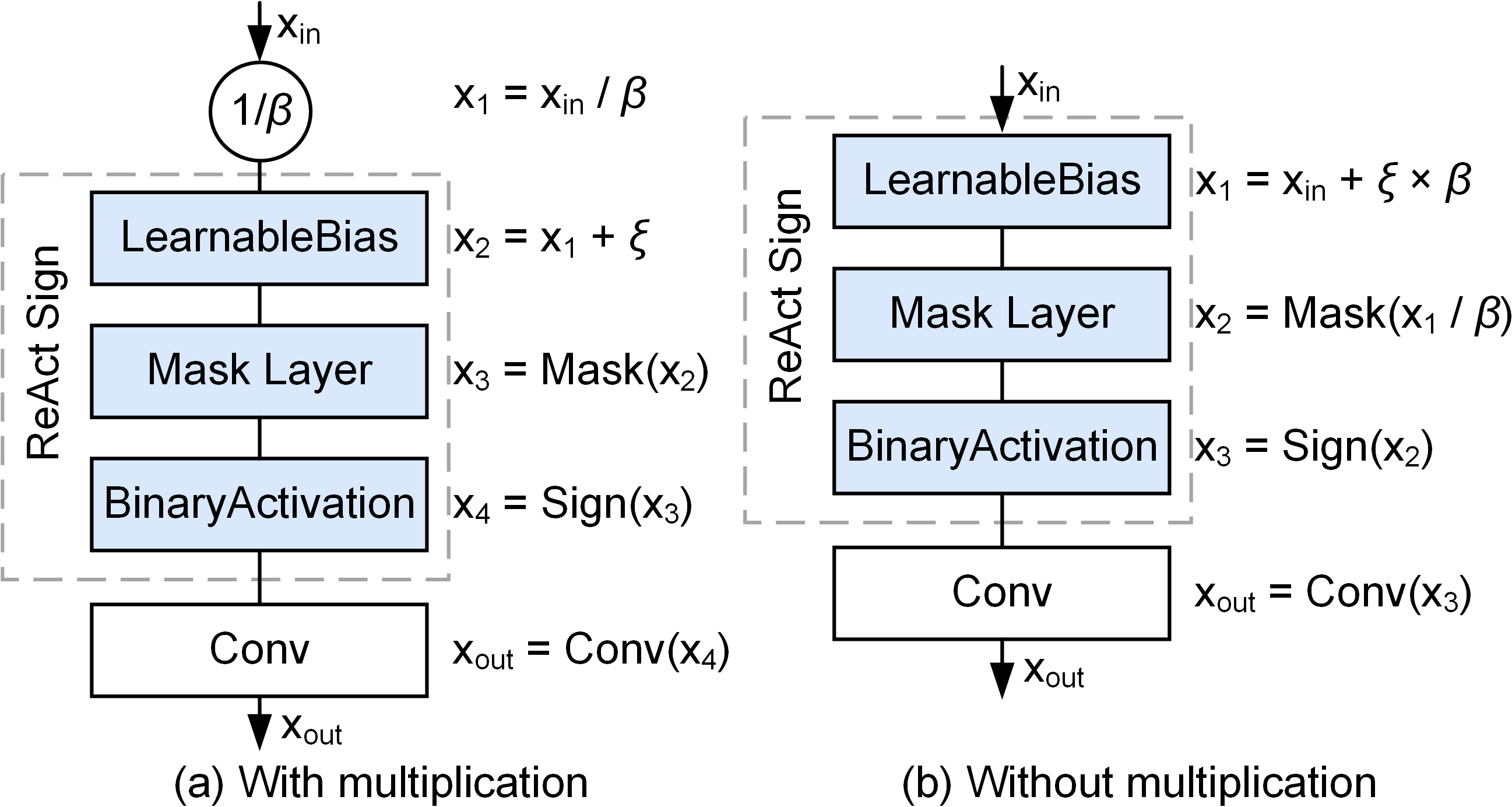}
\caption{(a) The original structure with multiplication. (b) The equivalent structure, although transforming one multiplication operation into two, can both be eliminated.}
\label{fig_MOFree}
\end{figure}

It is crucial to maintain the numerical symbol of the mask layer unchanged before and after mapping. This layer satisfies the relation:
\begin{equation}
\mathrm{Sign}(\mathrm{Mask}(x))=\mathrm{Sign}(\mathrm{Mask}(k\cdot x)) \equiv \mathrm{Sign}(x)
\end{equation}
where $k$ is a positive value. This relationship demonstrates the mathematical properties of the mask layer, enabling the absorption of a multiplication factor during forward propagation and effectively eliminating the multiplication operations caused by $\beta$. When backpropagating, it satisfies:
\begin{equation}
\frac{\partial y_{out}}{\partial x_{in}} = \frac{\partial y_{out}}{\partial y_{ML}} \frac{\partial y_{ML}}{\partial x_{in}} = 1 \cdot {f'}_A(x) \equiv {f'}_A(x)
\end{equation}
The same gradient transfer effect can be obtained. \Cref{fig_MOFree}a illustrates the network structure near $\beta$, showcasing the operation of each step.
In \cref{fig_MOFree}b, we propose an equivalent representation that equates one multiplication operation to two operations. At first glance, it might seem that instead of eliminating the multiplication operation, it increases it.
However, after completing network training, the product of $\xi$ and $\beta$ becomes a constant value that does not require recalculation.
Furthermore, by leveraging the characteristics of the mask layer, we can eliminate the second multiplication, achieving the same effect without any multiplications during inference.
The mathematical characteristics of the mask layer imply that it does not play a role during inference and can be directly removed.
It must be present during training as its significance lies in scaling and transforming values through mask function to regulate gradient transfer and prevent gradient vanishing.
We employ the mask function described in \cref{equ_4} and set $\delta=3$.
Its non-zero characteristic improves the activation saturation issue in BNN.
\begin{equation}
\mathrm{Sigmoid}(x,\delta) = \frac{1}{{1 + {e^{ - \delta x}}}}
\label{equ_4}
\end{equation}

%% --------------------------------------------------------------------------------------------------------------------------
\subsection{Bit Operations and Quantized RPRuLU Unit}

Using bit operations for multiplication by powers of 2 is a straightforward and efficient technique, which relies on the binary representation of numbers and the properties of shifting operations.
Multiplying a number by a power of 2 is equivalent to left-shifting its binary representation by a specific number of bits.
Shifting operations are highly optimized in many architectures and can be performed quickly, often surpassing traditional multiplication algorithms.
Within the BN-Free structure, the parameter $\alpha$ is typically assigned a small value, such as 0.2.
By conducting the parameter search, it is possible to set $\alpha$ to a negative integer power of 2, effectively substituting multiplication operations with an equal number of bit operations.
Similarly, all average pooling layer kernel sizes are set to $2\times2$, enabling the replacement of the division operation with a right shift of two bits.

\begin{figure}[!t]
\centering
\includegraphics[width=\linewidth]{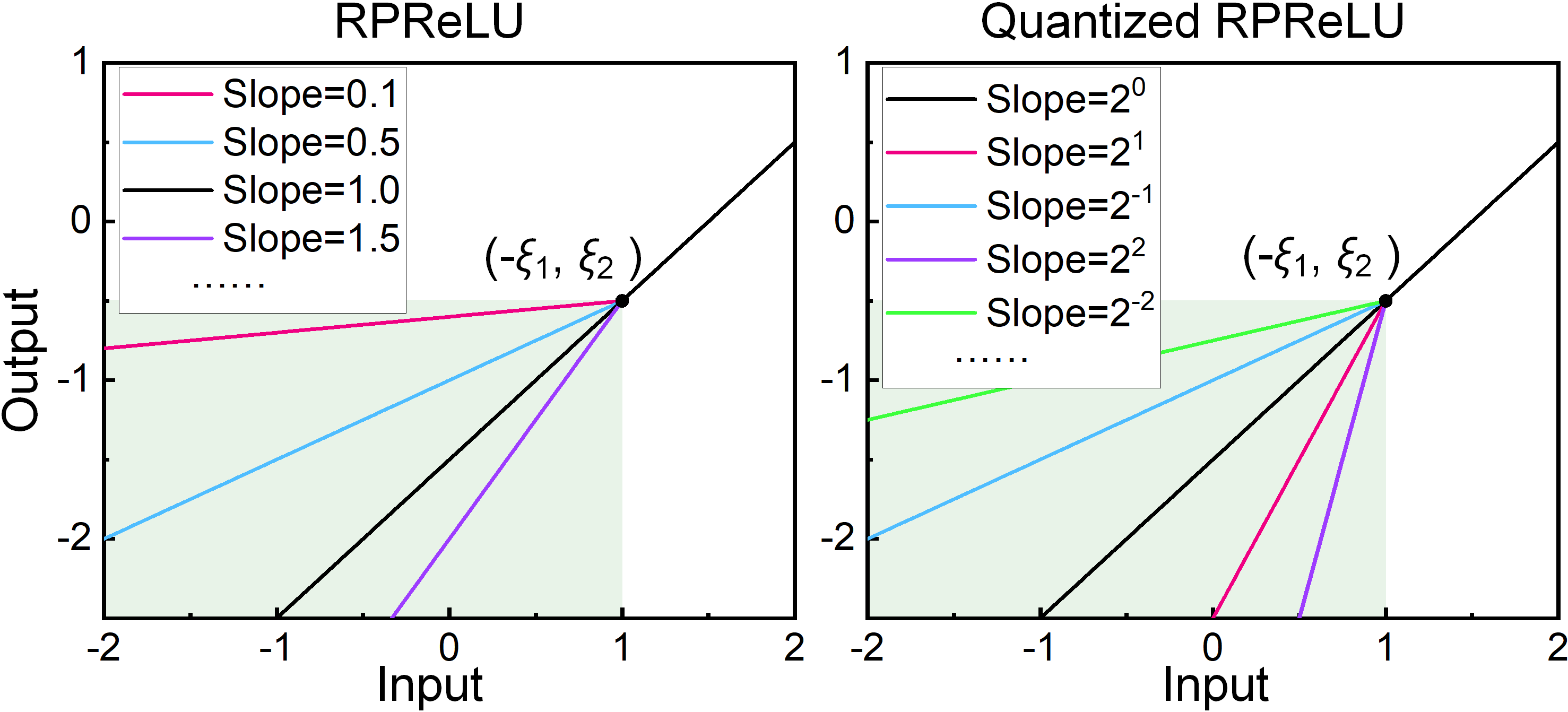}
\caption{The slope of RPReLU can be any continuous value greater than 0, while the slope of the proposed quantized RPReLU is only allowed to be an integer power of 2.}
\label{fig_prelu}
\end{figure}

When dealing with RPReLU, a straightforward approach is to substitute the PReLU with LeakyReLU and set the slope to a constant integer power of 2.
Nevertheless, the results of ablation studies indicate a decrease in performance with this approach.
To address this issue, we propose the quantized RPReLU unit.
In this unit, the parameters of PReLU's each channel are quantized values and are constrained to integer powers of 2, the expression is as follows:
\begin{equation}
\label{equ_6}
f(y_{i})=\left\{\begin{array}{ll}
y_{i}, & \text {if } y_{i} \ge 0 \\
2^{\mathrm{round}(a_{i})} \cdot (y_{i}+\xi_{i_1})+\xi_{i_2}, & \text{otherwise}
\end{array}\right.
\end{equation}
where $a_{i}$, $\xi_{i_1}$, and $\xi_{i_2}$ are all learnable parameters correspond to the channel $i$.
\Cref{fig_prelu} illustrates the function graph of PReLU and the proposed quantized RPReLU, respectively.
The green area represents the allowed slope range, which can take any continuous value in RPReLU, whereas only discrete quantized values are allowed in quantized RPReLU.
Ablation studies show that utilizing the quantized RPReLU can enhance accuracy by 1.14\% when compared to using RLeakyReLU on the ImageNet dataset.

%% --------------------------------------------------------------------------------------------------------------------------
\subsection{Hardware Benefits}
The A\&B BNN architecture proposed in this study holds significant practical importance.
We conducted synthesis on various units using the Xilinx Zynq-7000 Z-7045 FPGA, and the hardware overhead is presented in \cref{table_hardware}. The resources integrated on the chip include 218,600 Look-Up Tables (LUTs), 54,650 slices, and 900 Digital Signal Processors (DSPs).
The synthesis results demonstrate that the 32-bit full-precision multiplier consumes 47 LUTs, 12 slices, and 4 DSPs, whereas the bit-operator only necessitates 32 LUTs, 11 slices, and no DSP.
%This leads to a reduction in hardware overhead ranging from 0.33x to 1.94x. Similarly, the 32-bit bit-shift operator uses no DSP.
Standard PReLU structures demand 57 LUTs, 17 slices, and 4 DSPs, while quantized PReLU structures utilize only 32 LUTs, 9 slices, and no DSP.
We achieve this reduction by converting the multiplication operations introduced by the $\alpha$ parameter, the PRePU layer, and the average pooling layer into an equal number of bit-shift operations.
Additionally, we eliminate the multiplication operations associated with $\beta$.
Our approach does not introduce any additional computation or storage overhead.
For one convolution layer without pooling, the reduction in consumption for three hardware resources is 57.6\%, 51.2\%, and 100\%, respectively.
\begin{table}[!t]\small
\centering
\begin{tabular}{p{4.0em} p{5.4em}<{\centering} p{5.4em}<{\centering} p{5.4em}<{\centering}}
\toprule[1.5pt]
{Units} & {LUTs} & {Slice} & {DSPs} \\
\midrule[1pt]
{Multiplier} & {47} & {12} & {4} \\
{Bit-shift} & {32 ($\downarrow31.9\%$)} & {11 ($\downarrow8.3\%$)} & {0 ($\downarrow100\%$)} \\
\midrule[1pt]
{PReLU} & {57} & {17} & {4} \\
{QPReLU} & {32 ($\downarrow43.9\%$)} & {9 ($\downarrow47.1\%$)} & {0 ($\downarrow100\%$)} \\
\bottomrule[1.5pt]
\end{tabular}
\caption{Hardware overhead table for different units.}
\label{table_hardware}
\end{table}
%\begin{table}[!h]\small
%\centering
%\begin{tabular}{p{4.0em} p{3.0em}<{\centering} p{3.0em}<{\centering} p{3.0em}<{\centering}}
%\toprule[1.5pt]
%{Units} & {LUTs} & {Slice} & {DSPs} \\
%\midrule[1pt]
%{Adder} & {16} & {5} & {3} \\
%{Bit-shift} & {32} & {11} & {0} \\
%{Multiplier} & {47} & {12} & {4} \\
%\midrule[1pt]
%{PReLU} & {57} & {17} & {4} \\
%{QPReLU} & {32} & {9} & {0} \\
%\bottomrule[1.5pt]
%\end{tabular}
%\caption{Hardware overhead table for different units.}
%\label{table_hardware}
%\end{table}

In scenarios where the chip lacks built-in multiplication support, yet the neural network demands it, a viable solution entails transmitting intermediate results to the host for computation and subsequently sending them back to the chip.
This process results in frequent communication and introduces considerable delays.
In contrast, the A\&B BNN architecture enables the network to perform inference entirely within the chip, necessitating only a single round-trip communication.
This streamlined approach significantly diminishes latency and improves real-time performance.

%% =============================================================== Experiments =============================================================== %% 
\section{Experiments}
\label{sec:Experiments}

%% --------------------------------------------------------------------------------------------------------------------------
\subsection{Experimental Setup}

We conducted experiments on the CIFAR-10 \cite{krizhevsky2009learning}, CIFAR-100 \cite{krizhevsky2009learning}, and ILSVRC12 ImageNet \cite{russakovsky2015imagenet} datasets using the advanced ReActNet-18/34 and ReActNet-A network structures in binary networks and presented the results.
In line with most binarization works, the input and output layers' weights are not quantized to ensure performance.
All seeds were fixed to 2023 to ensure the experiments' repeatability.

{\bf{Implementation details for ImageNet.}}
ImageNet is a widely used benchmark dataset in computer vision, consisting of over 1.28 million training images, 50,000 test images, and 1,000 classes.
Due to the significant success of ReActNet in the binary domain, we utilized our add\&bit-operation-only network on ReActNet-18/34 and ReActNet-A, which are enhanced versions of ResNet-18/34 \cite{he2016deep} and MobileNetv1 \cite{howard2017mobilenets} structures, respectively.
Additionally, we employed SWS and AGC techniques, setting $\lambda$ to 0.02 for stability based on \cite{chen2021bnn}.
For training, we employed a two-step strategy \cite{martinez2019training}, i.e., initially training from scratch and binarizing only the activations in the first stage, then fine-tuning based on the previous stage and binarizing both weights and activations in the second stage.
At each stage, we utilized the Adam optimizer to minimize the Distillation loss functions \cite{liu2020reactnet} and training for 128 epochs, starting with an initial learning rate of 1e-3 and gradually decreasing to zero using a linear scheduler.
The weight decay is set to 5e-6 in the first stage and 0 in the second stage.
We use the same data augmentation as \cite{chen2021bnn, liu2020reactnet}, including random cropping, lighting, and random horizontal flipping, and use the same knowledge distillation scheme.
For the test set, the image is resized to 256, center-cropped to 224, and then inputted into the network.

{\bf{Implementation details for CIFAR-10 and CIFAR-100.}}
CIFAR-10 and CIFAR-100 datasets consist of 50,000 training images and 10,000 testing images, divided into 10 and 100 classes, respectively.
We conducted experiments using the ReActNet-18 and ReActNet-A structures, following the same two-step training strategy as in the ImageNet experiments, with each step trained for 256 epochs.
To illustrate the versatility of the proposed algorithm, experiments were also conducted on other two mainstream binary structures Bi-real BNN and XNOR BNN.
Since the ReActNet-A architecture includes more sub-sampling units and is not well-suited for handling small-sized datasets such as CIFAR, we adjusted the stride of the initial layer to 1 like ResNet.
Data augmentation included random cropping and horizontal flipping.
Additionally, we adopted the SWS and AGC techniques, setting $\lambda$ to 0.001 based on \cite{chen2021bnn}.
The remaining experimental settings align with those used in the ImageNet experiments.

%% --------------------------------------------------------------------------------------------------------------------------
\subsection{Comparison with State-of-the-Arts}

\begin{figure}[!t]
\centering
\includegraphics[width=\linewidth]{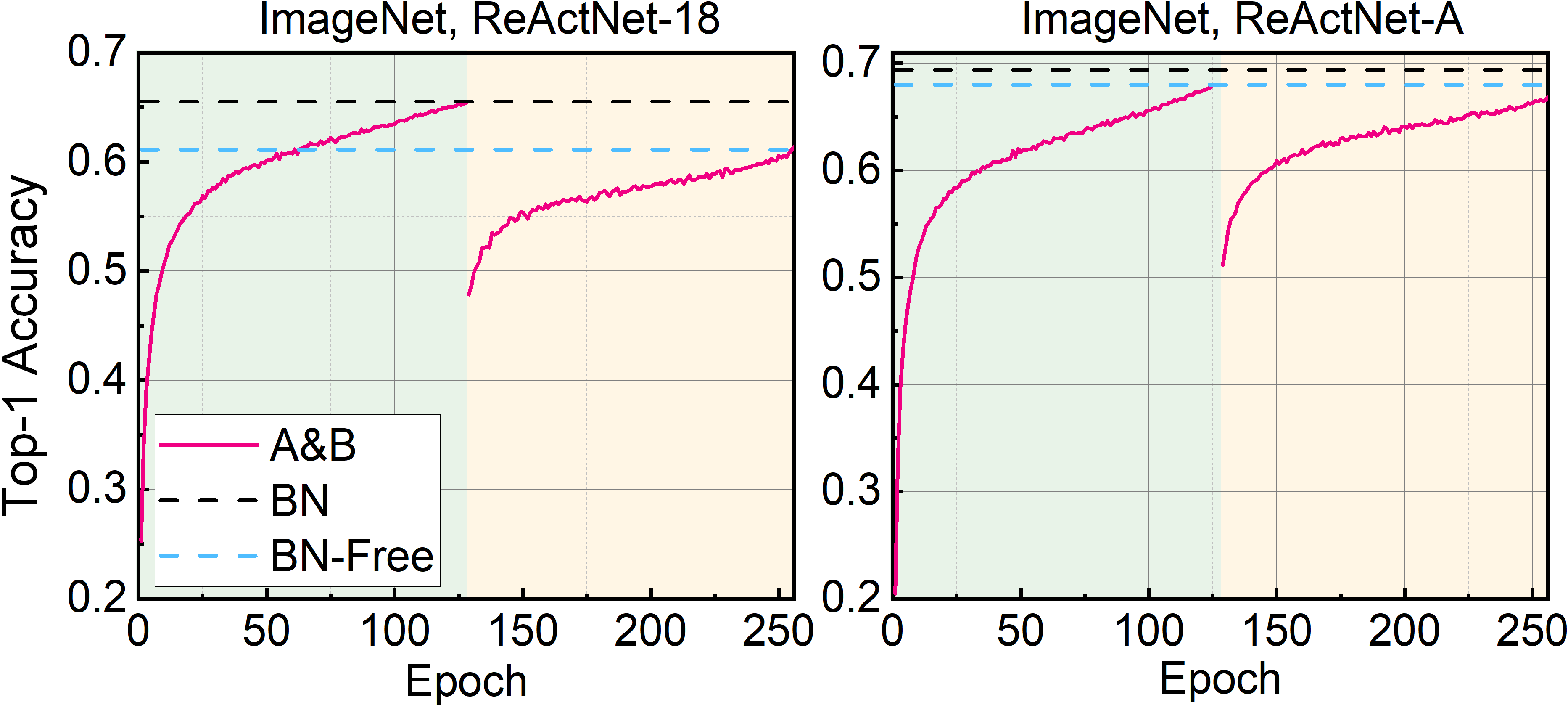}
\caption{Top-1 accuracy on the ImageNet dataset, and comparison with two baselines. The green background represents the first training step and the orange represents the second step.}
\label{fig_ImageNet}
\end{figure}

{\bf{Results on ImageNet.}}
We first tested the performance of the proposed A\&B ReActNet-18/34 and ReActNet-A models on ImageNet to validate the algorithm's effectiveness.
The Top-1 accuracies for the three network structures were 61.39\%, 65.19\%, and 66.89\%, while the Top-5 accuracies were 83.06\%, 86.03\%, and 86.83\%, respectively.
The comparison results are presented in \cref{table_compare}, and \cref{fig_ImageNet} illustrates the Top-1 training results.
Compared to the BN-Free structure, we achieved a reduction of 14.7 million multiplication operations with an accuracy loss of 1.11\% on ReActNet-A.
We achieved a reduction of 4.6 million multiplication operations on ReActNet-18 while improving the accuracy by 0.29\%.
Considering the actual application scenarios of BNN, its model complexity is much lower than MobileNet, and the task difficulty is much lower than ImageNet.
The only 1.1\% performance loss on ImageNet is difficult to feel on edge applications.
Conversely, the hardware overhead is an urgent problem to be solved, the multiplier-less structure proposed is of great significance in actual production \cite{simons2019review}.
%There is a great temptation in practical applications to replace all multiplication operations while maintaining an acceptable performance loss.
%\begin{table}[!h]
%\caption{Accuracy on the ImageNet Dataset with Two Baselines.\label{table_ImageNet}}
%\centering
%\begin{tabular}{p{11.7em} p{2.4em}<{\centering} p{2.4em}<{\centering}}
%\toprule[1.5pt]
%\multirow{3}{*}{Binary Network} &  \multicolumn{2}{c}{Acc (\%)}\\
%\cmidrule{2-3}
% & Top-1 & Top-5 \\
%\midrule[1pt]
%ReActNet-18 \cite{liu2020reactnet} & 65.5 & - \\
%ReActNet-18 (BN-Free) \cite{chen2021bnn} & 61.1 & - \\
%\textbf{ReActNet-18 (MO-Free)} & \textbf{61.39} & \textbf{83.06} \\
%\midrule[1pt]
%ReActNet-A \cite{liu2020reactnet} & 69.4 & - \\
%ReActNet-A (BN-Free) \cite{chen2021bnn} & 68.0 & - \\
%\textbf{ReActNet-A (MO-Free)} & \textbf{66.89} & \textbf{86.83} \\
%\bottomrule[1.5pt]
%\end{tabular}
%\end{table}
\begin{figure*}[!t]
\centering
\includegraphics[width=6.5in]{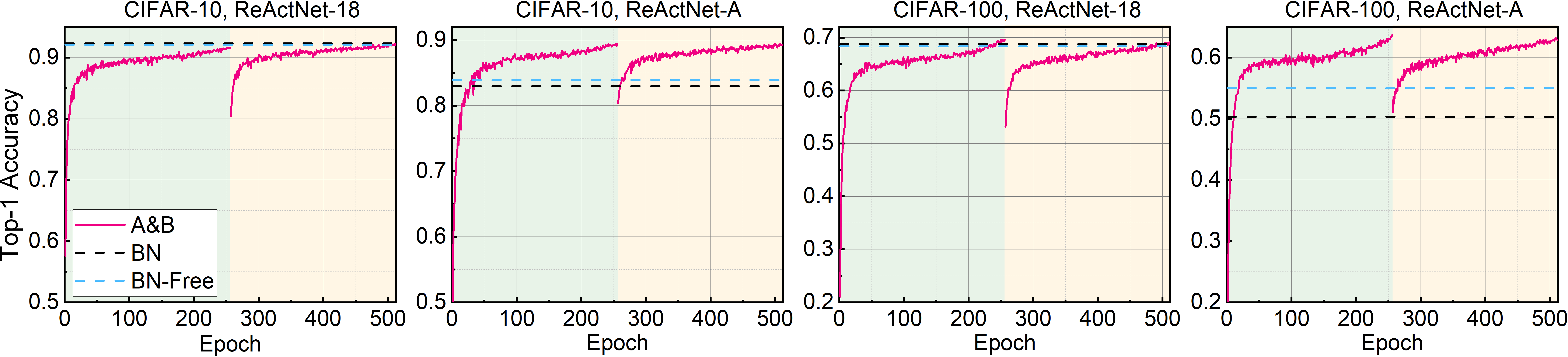}
\caption{Top-1 accuracy on the CIFAR-10 and CIFAR-100 datasets, and comparison with two baselines.}
\label{fig_CIFAR}
\end{figure*}

{\bf{Results on CIFAR-10 and CIFAR-100.}}
To further validate the algorithm's effectiveness and versatility, we conducted additional experiments using the widely used smaller datasets CIFAR-10 and CIFAR-100.
The comparison results are presented in \cref{table_CIFAR}, while the Top-1 training results are depicted in \cref{fig_CIFAR}.
The algorithm demonstrated competitive results on both datasets.
The results of ReActNet-A on CIFAR validate the intuition of modifying the structure.
Merely reducing the stride of the first layer convolution can lead to a substantial improvement ranging from 5.53\% to 12.93\%.
Since quantization methods often have higher variance, providing the mean and standard deviation would provide a better understanding of the experimental results.
Experiments with ReActNet-18 on CIFAR10/100 using five adjacent seeds were conducted, achieving $\mu=92.31\%/69.37\%$, $\sigma=6.35e-4/5.40e-4$.
%\begin{table}[!h]\small
%\centering
%\begin{tabular}{p{11.6em} p{4.3em}<{\centering} p{4.8em}<{\centering}}
%\toprule[1.5pt]
%\multirow{3}{*}{Binary Network} &  \multicolumn{2}{c}{Acc}\\
%\cmidrule{2-3}
% & CIFAR-10 & CIFAR-100 \\
%\midrule[1pt]
%ReActNet-18 \cite{liu2020reactnet} & 92.31\% & 68.78\% \\
%ReActNet-18 (BN-Free) \cite{chen2021bnn} & 92.08\% & 68.34\% \\
%\textbf{ReActNet-18 (A\&B BNN)} & \textbf{92.30\%} & \textbf{69.35\%} \\
%\midrule[1pt]
%ReActNet-A \cite{liu2020reactnet} & 82.95\% & 50.30\% \\
%ReActNet-A (BN-Free) \cite{chen2021bnn} & 83.91\% & 55.00\% \\
%\textbf{ReActNet-A (A\&B BNN)} & \textbf{89.44\%} & \textbf{63.23\%} \\
%\bottomrule[1.5pt]
%\end{tabular}
%\caption{Accuracy on the CIFAR-10 and CIFAR-100 datasets with two baselines.}
%\label{table_CIFAR}
%\end{table}
\begin{table}[!t]\footnotesize
\centering
%\scriptsize % 设置表格内字体大小为小号
\begin{tabular}{p{4.6em}<{\centering} p{1.9em}<{\centering} p{3.5em}<{\centering} p{1.9em}<{\centering} | p{1.9em}<{\centering} p{3.5em}<{\centering} p{1.9em}<{\centering}}
\toprule[1.5pt]
\multirow{3}{*}{BNNs} & \multicolumn{3}{c}{CIFAR10 (\%)} & \multicolumn{3}{c}{CIFAR100 (\%)} \\
\cmidrule{2-7}
 & BN & BN-Free & A\&B & BN & BN-Free & A\&B \\
\midrule[1pt]
XnorNet18 & {90.21} & {79.67} & {89.94} & {65.35} & {53.76} & {64.51} \\
\cmidrule{1-7}
BiResNet18 & {89.12} & {79.59} & {90.09} & {63.51} & {54.34} & {65.52} \\
\cmidrule{1-7}
ReActNet18 & {92.31} & {92.08} & {92.30} & {68.78} & {68.34} & {69.35} \\
\cmidrule{1-7}
ReActNetA & {82.95} & {83.91} & {89.44} & {50.30} & {55.00} & {63.23} \\
\bottomrule[1.5pt]
\end{tabular}
\caption{Accuracy on the CIFAR-10 and CIFAR-100 datasets with two baselines.}
\label{table_CIFAR}
\end{table}

%% --------------------------------------------------------------------------------------------------------------------------
\subsection{Ablation Study}
In the previous section, we demonstrated the effectiveness of the proposed A\&B BNN and achieved competitive results.
In this section, we further illustrate the necessity of the proposed quantized RPReLU and its impact on network performance through ablation studies and then explore the impact of the hyperparameter $\alpha$.
Due to computational constraints, all ablation studies are conducted using the ReActNet-18 structure.

\begin{figure}[!t]
\centering
\includegraphics[width=\linewidth]{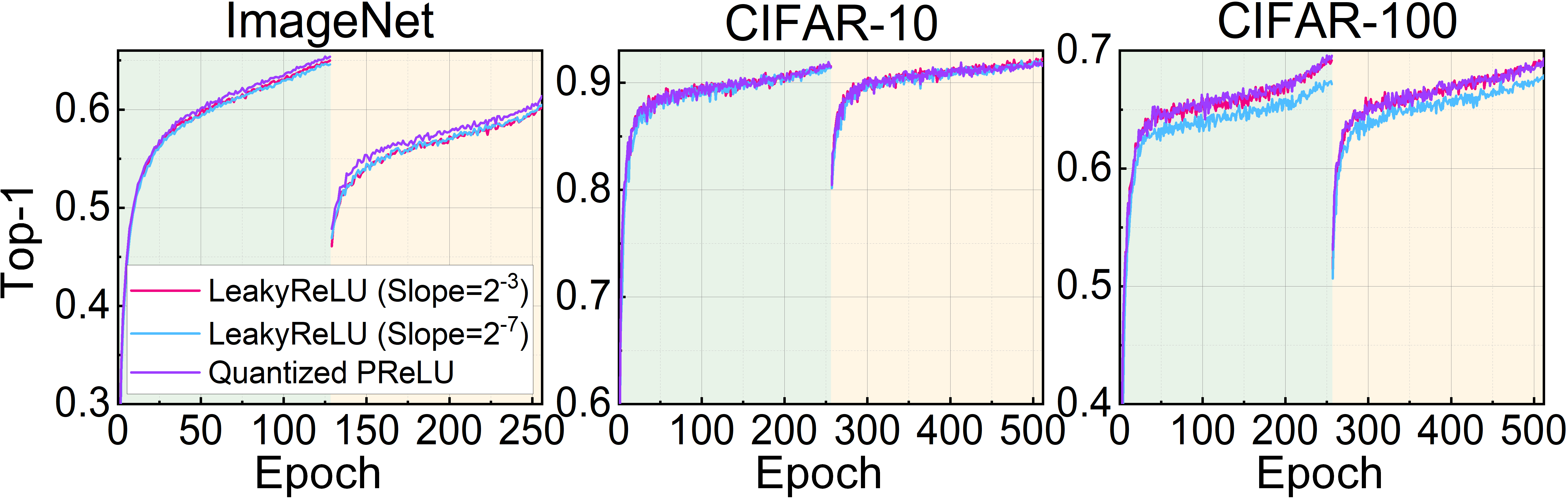}
\caption{Ablation studies results of different ReLU structures. The findings demonstrate that employing the proposed quantized RPReLU architecture enhances performance on the ImageNet dataset by 1.14\% when compared to the fixed-value LeakyReLU.}
\label{fig_a1}
\end{figure}

{\bf{Quantized RPReLU unit.}}
To eliminate the multiplication operation introduced by RPReLU, we propose replacing PReLU with LeakyReLU or utilizing quantized PReLU.
Although LeakyReLU with a fixed slope is a straightforward option, it can lead to performance degradation on complex datasets. To compare the effectiveness of different options, \cref{table_a1} provides a comparison between quantized RPReLU and RLeakyReLU with a fixed slope of $2^{-3}$/$2^{-7}$, which approximates the commonly used slopes of 0.1 and 0.01. \Cref{fig_a1} depicts the Top-1 accuracies conducted on three datasets. The results exhibit a significant disparity in the impact of the activation function type between the simple dataset CIFAR and the more complex dataset ImageNet. In the case of the ImageNet dataset, the activation function type has only a 0.40\% impact during the first stage of activations binarization, but the second-stage impact on the binarization of both activations and weights reaches 1.14\%, demonstrating the contribution of activation function nonlinearity in enhancing network performance. The complexity of the convolution structure with both weight and activation binarized is substantially decreased. Weak nonlinearity in the activation function can lead to a deterioration in multi-layer convolutions, reducing the network's capacity to accommodate the data. Although it introduces only discrete integer powers of 2, quantized PReLU leads to a substantial improvement in activation function nonlinearity and enhances network expression. The experimental results based on the ReActNet-A structure in \cref{table_compare} employing BN-Free and A\&B BNN also offer supporting evidence for this claim.
\begin{table}[!t]\footnotesize
\centering
%\scriptsize % 设置表格内字体大小为小号
\begin{tabular}{p{4.8em}<{\centering} p{2.3em}<{\centering} p{2.3em}<{\centering} | p{2.3em}<{\centering} p{2.3em}<{\centering} | p{2.3em}<{\centering} p{2.3em}<{\centering}}
\toprule[1.5pt]
\multirow{3}{*}{Type} & \multicolumn{2}{c}{ImageNet (\%)} & \multicolumn{2}{c}{CIFAR-10 (\%)} & \multicolumn{2}{c}{CIFAR-100 (\%)} \\
\cmidrule{2-7}
 & Step1 & Step2 & Step1 & Step2 & Step1 & Step2 \\
\midrule[1pt]
RLeakyReLU & \multirow{2}{*}{65.00} & \multirow{2}{*}{60.25} & \multirow{2}{*}{91.67} & \multirow{2}{*}{\textbf{92.30}} & \multirow{2}{*}{69.22} & \multirow{2}{*}{69.31} \\
(Slope=$2^{-3}$) & & & & & & \\
\cmidrule{1-7}
RLeakyReLU & \multirow{2}{*}{64.66} & \multirow{2}{*}{60.48} & \multirow{2}{*}{91.61} & \multirow{2}{*}{92.07} & \multirow{2}{*}{67.42} & \multirow{2}{*}{67.90} \\
(Slope=$2^{-7}$) & & & & & & \\
\cmidrule{1-7}
Quantized & \multirow{2}{*}{\textbf{65.40}} & \multirow{2}{*}{\textbf{61.39}} & \multirow{2}{*}{\textbf{91.94}} & \multirow{2}{*}{91.94} & \multirow{2}{*}{\textbf{69.59}} & \multirow{2}{*}{\textbf{69.35}} \\
RPReLU & & & & & & \\
\bottomrule[1.5pt]
\end{tabular}
\caption{Ablation studies results of different ReLU structures.}
\label{table_a1}
\end{table}
The sole difference in network expression ability between the two resides in whether PReLU is quantized, leading to a 1.11\% improvement, which further highlights the significance of nonlinearity in ReActNet-A on ImageNet.
For ReActNet-18, whether RPReLU is quantized has little effect on the results, which also applies to the small dataset CIFAR.
This implies that network performance in this case is influenced by network complexity, rather than nonlinearity.

\begin{figure}[!t]
\centering
\includegraphics[width=\linewidth]{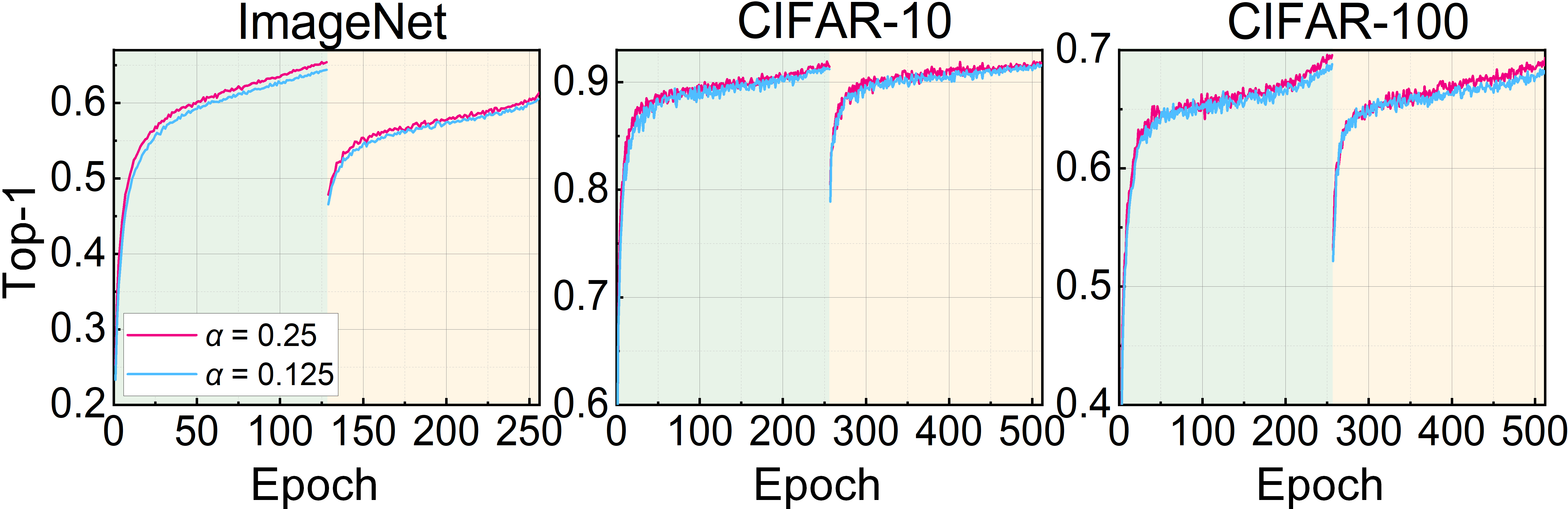}
\caption{Ablation studies results using different $\alpha$ values. The results show that the value of 0.25 is consistently better than the value of 0.125.}
\label{fig_a2}
\end{figure}

{\bf{Hyperparameter $\alpha$.}}
The factor $\alpha$ is utilized after the convolutional layer to provide feedback to RPReLU, typically set to a small value such as 0.2.
In \cite{brock2020characterizing, brock2021high}, $\alpha$ is used to control the growth rate of the variance of the residual structure, and a large $\alpha$ brings fast growth while a small $\alpha$ will weaken the effect of the residual \cite{he2016identity}, which requires a trade-off.
To explore the impact of $\alpha$'s value on the network's performance, we set it to the closest quantized values to 0.2, specifically $2^{-2}$ and $2^{-3}$.
The results are presented in \cref{table_a2} and \cref{fig_a2} shows the results of Top-1 ablation studies conducted on three datasets.
The experimental results consistently demonstrate that employing $\alpha=2^{-2}$ yields better effects, with the accuracy increasing by 0.25\% to 1\% compared to $\alpha=2^{-3}$.
\begin{table}[!t]\footnotesize
\centering
%\scriptsize % 设置表格内字体大小为小号
\begin{tabular}{p{2.6em}<{\centering} p{2.3em}<{\centering} p{2.3em}<{\centering} | p{2.3em}<{\centering} p{2.3em}<{\centering} | p{2.3em}<{\centering} p{2.3em}<{\centering}}
\toprule[1.5pt]
\multirow{3}{*}{$\alpha$} & \multicolumn{2}{c}{ImageNet (\%)} & \multicolumn{2}{c}{CIFAR-10 (\%)} & \multicolumn{2}{c}{CIFAR-100 (\%)} \\
\cmidrule{2-7}
 & Step1 & Step2 & Step1 & Step2 & Step1 & Step2 \\
\midrule[1pt]
$2^{-2}$ & \textbf{65.40} & \textbf{61.39} & \textbf{91.94} & \textbf{91.94} & \textbf{69.59} & \textbf{69.35} \\
\cmidrule{1-7}
$2^{-3}$ & 64.38 & 60.55 & 91.38 & 91.69 & 68.81 & 68.35\\
\bottomrule[1.5pt]
\end{tabular}
\caption{Ablation studies results of different $\alpha$ values.}
\label{table_a2}
\end{table}

\subsection{Visualization}
\begin{figure}[!t]
\centering
\includegraphics[width=\linewidth]{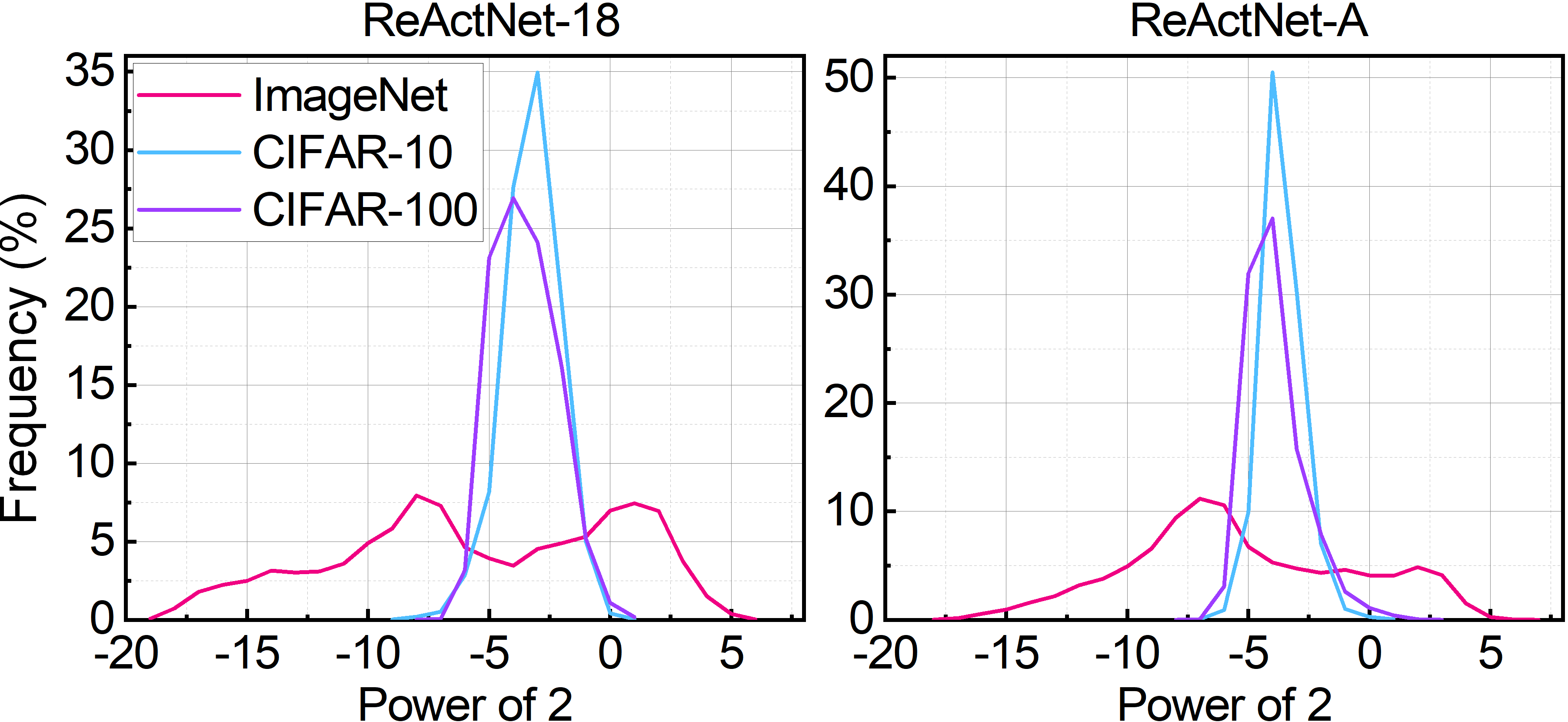}
\caption{Visualization of the quantified RPReLU architecture shows the distribution of slopes for ReActNet-18 and ReActNet-A structures following training on CIFAR-10, CIFAR-100, and ImageNet datasets.}
\label{fig_v1}
\end{figure}

The effectiveness of the proposed quantized RPReLU and its ability to enhance network nonlinearity are demonstrated through ablation studies. This section presents visualizations of the range of quantized values and their frequencies in the quantized RPReLU. \Cref{fig_v1} illustrates the distribution of quantization slopes for the trained network, which quantizes RPReLU on three datasets using ReActNet-18 and ReActNet-A, respectively. The figure reveals that the quantization values are mainly distributed within the range of $2^{-19}$ to $2^{6}$. For the small dataset CIFAR, the exponential distribution ranges from -7 to 2, with a concentration around -3. For the ImageNet dataset, the quantization distribution is wider, ranging from -19 to 7, with approximately two peaks centered around -6 and 2. A broader and more balanced distribution is more representative of non-linearities.

%% =============================================================== Conclusion =============================================================== %% 
\section{Conclusion}
\label{sec:Conclusion}

This study introduces a novel binary network architecture, A\&B BNN, designed to perform network inference without any multiplication operation. Building upon the BN-Free architecture, we introduced the mask layer and the quantized RPReLU structure to completely remove all multiplication operations within the traditional binary network. The mask layer leverages the inherent features of BNNs and employs straightforward mathematical transformations, allowing for its direct removal during inference to eliminate associated multiplication operations. Additionally, the quantized RPReLU structure enhances efficiency by constraining its slope to an integer power of 2, employing bit operations rather than multiplications. Experimental results show that A\&B BNN achieved accuracies of 92.30\%, 69.35\%, and 66.89\% on CIFAR-10, CIFAR-100, and ImageNet datasets, respectively, which is competitive with state-of-the-art. This study introduces a novel insight for creating hardware-friendly binary neural networks.

\section{Acknowledgment}
This work was supported by STI 2030-Major Projects 2022ZD0209700.

{
    \small
    \bibliographystyle{ieeenat_fullname}
    \bibliography{main}
}

% WARNING: do not forget to delete the supplementary pages from your submission 
%\input{sec/X_suppl}

\end{document}